\documentclass[sigconf]{acmart}
\setcopyright{none}
\renewcommand\footnotetextcopyrightpermission[1]{}
\setcopyright{none}
\makeatletter
\renewcommand\@formatdoi[1]{\ignorespaces}
\makeatother   
   \usepackage{balance}
\usepackage{times}
\usepackage{amsmath}
\usepackage{amsthm}
\usepackage{amssymb}
\usepackage{multirow}
\usepackage{mathtools}

\def\BibTeX{{\rm B\kern-.05em{\sc i\kern-.025em b}\kern-.08emT\kern-.1667em\lower.7ex\hbox{E}\kern-.125emX}}

\begin{document}

\fancyhead{}
  % do not delete this code.

\title{Compact Network Training for Person ReID}

% The "author" command and its associated commands are used to define the authors and their affiliations.
% Of note is the shared affiliation of the first two authors, and the "authornote" and "authornotemark" commands
% used to denote shared contribution to the research.
% {\tt\small $\{$hussam.lawen, avi.bencohen, matan.protter, itamar.friedman, lihi.zelnik $\}$} \\ {\tt\small @alibaba-inc.com }
% }
\author{Hussam Lawen}
\authornote{Both authors contributed equally to this research.}
\email{hussam.lawen@alibaba-inc.com}

\author{Avi Ben-Cohen}
\authornotemark[1]
\email{avi.bencohen@alibaba-inc.com}
\affiliation{
  \institution{DAMO Academy, Alibaba Group}      
  \city{Tel-Aviv}
  \country{Israel}}

  \author{Matan Protter}
  \email{matan.protter@alibaba-inc.com}
  \affiliation{
    \institution{DAMO Academy, Alibaba Group}
    \city{Tel-Aviv}
  \country{Israel}}

    \author{Itamar Friedman}
    \email{itamar.friedman@alibaba-inc.com}
    \affiliation{
      \institution{DAMO Academy, Alibaba Group}      
      \city{Tel-Aviv}
  \country{Israel}}

      \author{Lihi Zelnik-Manor}
      \email{lihi.zelnik@alibaba-inc.com}
      \affiliation{
        \institution{DAMO Academy, Alibaba Group}      
        \city{Tel-Aviv}
  \country{Israel}}

\renewcommand{\shortauthors}{Lawen and Ben-Cohen, et al.}

\begin{abstract}
   The task of person re-identification (ReID) has attracted growing attention in recent years leading to improved performance, albeit with little focus on real-world applications.
   Most SotA methods are based on heavy pre-trained models, e.g. ResNet50 ({\raise.17ex\hbox{$\scriptstyle\sim$}}25M parameters), which makes them less practical and more tedious to explore architecture modifications.
In this study, we focus on a small-sized randomly initialized model that enables us to easily introduce architecture and training modifications suitable for person ReID.
The outcomes of our study are a compact network and a fitting training regime.
   We show the robustness of the network by outperforming the SotA on both Market1501 and DukeMTMC. Furthermore, we show the representation power of our ReID network via SotA results on a different task of multi-object tracking.
\end{abstract}



\maketitle

\section{Introduction}

The objective in person re-identification (ReID) is to assign a stable ID to a person in multiple camera views.
In this study we are interested in the development of small sized models for ReID with high accuracy for two main reasons. First, it is beneficial for practical deployment and productization of ReID solutions.
Second, the research for models that provide high accuracy requires exploration of many architecture variations and training schemes.
When the backbone is heavy, re-training consumes both a lot of time and computing resources which we wish to avoid.
Our approach differs from many state-of-the-art (SotA) methods, that rely on large pre-trained backbone models, such as ResNet50, e.g.~\cite{wu2019deep,sun2018pcb,wang2018mancs,luo2019bag}.

\begin{figure}
\centering
\begin{minipage}[b]{0.45\textwidth}
\includegraphics[width=1\textwidth]{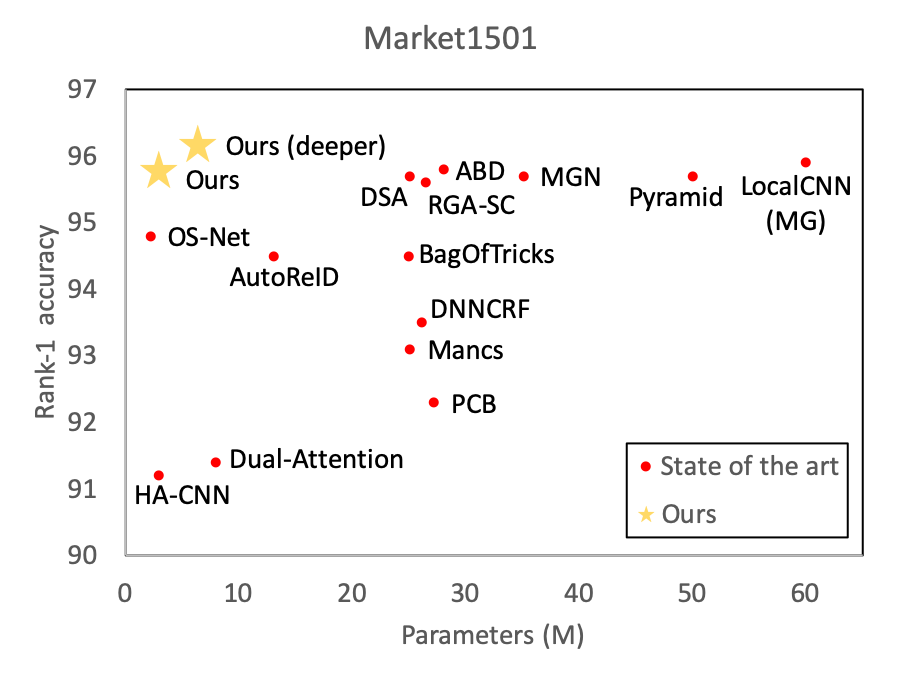}
\end{minipage}

\centering
\begin{minipage}[b]{0.45\textwidth}
\includegraphics[width=1\textwidth]{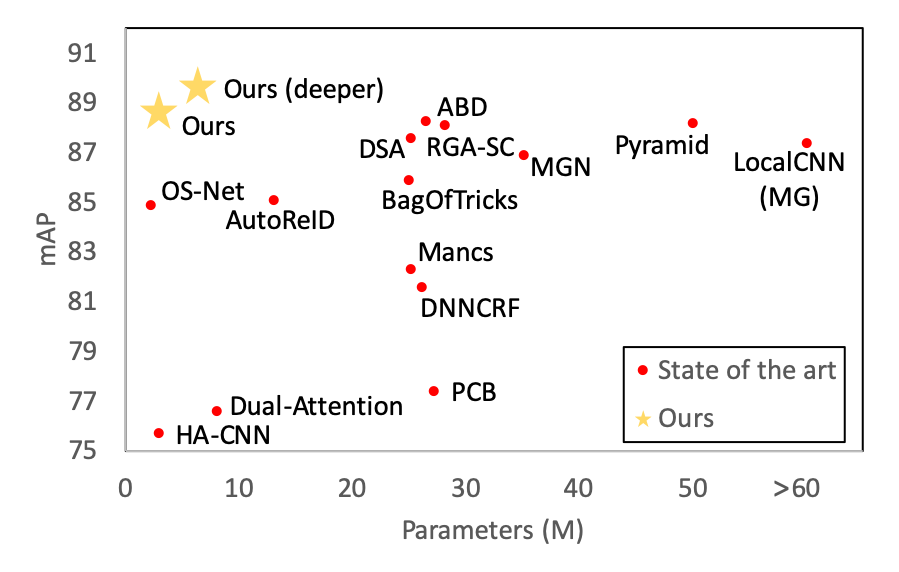}
\end{minipage}
\caption{Performance comparison of our approach and SotA ReID methods on Market1501 dataset. Top: rank-1 accuracy vs. number of parameters. Bottom: mAP vs. number of parameters.}\label{fig:param_market}
\end{figure}

We argue that a cost-effective ReID model should be computationally efficient, capable of running on low-res video input, and robust to multiple camera setting.
Hence, we propose an efficient ReID model and training schemes that demonstrate state of the art performance under these requirements.
To reduce the computational burden, we aim to decrease the number of parameters and use a relatively small ReID model. Figure \ref{fig:param_market} shows the current state of the art results \cite{chen2018dnncrf,li2018harmonious,luo2019bag,quan2019auto,si2018dual,sun2018pcb,wang2018mancs,yang2018local,zheng2018pyramid,zhang2019densely,zhang2019relation,wang2018learning,chen2019abd,zhou2019omni} and the number of parameters compared to our proposed method on the popular Market1501 dataset \cite{zheng2015market} in terms of rank-1 accuracy and mAP. For some methods, the number of parameters was not known so we used an estimated lower bound. Using our proposed training framework we achieve state of the art results with an order of magnitude smaller model compared to the best existing ReID CNN.

The importance of training ``tricks'' for deep person ReID has been discussed before in~\cite{luo2019bag}.
% In this paper, we suggest additional training techniques and architecture modifications that can further improve the model performance and achieve similar or better results than much larger and complicated models on several datasets.
In this paper, we suggest training techniques and architecture modifications that improve the harmonious attention network HA-CNN of~\cite{li2018harmonious} to achieve similar or better results than much larger and complicated models.
The contribution of this paper is thus three-fold:
\begin{itemize}
  \item  We propose a compact and robust deep person ReID model. Our model achieves state of the art results on two popular person ReID datasets (Market1501 and DukeMTMC ReID \cite{ristani2016dukemtmc}). This is despite having a small number of parameters, small number of FLOPS, and low resolution input image, in comparison to other leading methods.

  \item We study a variety of training schemes and network choices that prove useful. While we have explored their affect only for HA-CNN, we believe they could be of interest for others to examine in other setups.

  \item We demonstrate the utility of the proposed person ReID model also for other tasks, by improving multi-target multi-camera tracking.

\end{itemize}

In the following section we describe the baseline ReID network we started with. The training techniques and architecture modifications that were explored in this study are presented in section \ref{sec:tricks}. Next, the experimental results including an ablation study, additional analysis, and comparison to state of the art are presented (section \ref{sec:experiments}). Finally, multi camera multi target tracking results are presented in section \ref{sec:tracking}.

\section{Baseline ReID network - HA-CNN}
Our goal is a compact model that gives high accuracy with low-resolution input images, in order to reduce computational complexity.
Therefore, we chose as baseline the light-weight Harmonious Attention CNN (HA-CNN)~\cite{li2018harmonious}. HA-CNN is sufficiently compact to be trained from scratch thus obviating the need to pre-train on additional data. Nonetheless, it provides good results taking into consideration its small number of parameters (2.7M).
In addition, the input image size for this network is relatively small compared to other person ReID networks.

HA-CNN is an attention network with several attention modules including soft spatial and channel-wise attention and hard attention to extract local regions. The network architecture holds two branches: a global one, and a local one that uses the regions extracted based on the hard attention. Finally, the output vectors of both branches are concatenated for the final person image descriptor. Holding two branches and multiple attention modules improves the network perception, despite these features the HA-CNN keeps a small number of parameters making it accurate and efficient. However, parts of the architecture can still be optimized as well as the training scheme. Optimizing it can further improve the HA-CNN and obtain a more accurate model.

\section{Methods}
\label{sec:tricks}
It is well known that the performance of deep learning models is highly dependent on both the choice of architecture and the training scheme. Specifically, recent work has shown that training procedure refinements can significantly improve ReID results~\cite{luo2019bag}.
In the following we explore training schemes (Section \ref{sec:training_techniques}) and architecture modifications (section \ref{sec:architecture_modifications}) that lead to better ReID performance of HA-CNN.
To make our survey more complete we further mention several modifications that did not improve the model performance (Section~\ref{sec:tried}).

\subsection{Training techniques}
\label{sec:training_techniques}
The following training techniques were found useful by our study:

\paragraph{Weighted triplet loss with Soft margin}
The triplet loss is widely used to train Person ReID models, as well as other computer vision tasks such as Face Recognition and Few-Shot Learning. The original triplet loss was proposed by Schroff \textit{et al.} \cite{schroff2015facenet}. We denote an anchor sample by $x_a$, positive samples as $x_p \in P(a)$ and negative samples as $x_n \in N(a)$, then the triplet loss can be written as:

\begin{equation}
L_{1} = \left[ m +\hspace{1pt} d(x_a,x_p) - \hspace{1pt} d(x_a,x_n) \right]_+
\label{eq:triplet_general}
\end{equation}
where $m$ is the given inter-class separation margin, $d$ denotes distance of appearance, and $[\boldsymbol{\cdot}]_+ = max(0,\boldsymbol{\cdot})$.

Hermans \textit{et al.} \cite{hermans2017defense} proposed the batch-hard triplet loss that selects only the most difficult positive and negative samples:
\begin{equation}
L_{2} = \left[ m +\max\limits_{\mathclap{\substack{x_p \in P(a)}}} \hspace{1pt} d(x_a,x_p) - \min\limits_{\mathclap{\substack{x_n \in N(a)}}} \hspace{1pt} d(x_a,x_n) \right]_+
\label{eq:hard_triplet}
\end{equation}
In contrast to the original triplet loss, the batch-hard triplet loss emphasizes hard examples. However, it is sensitive to outlier samples and may discard useful information due to its hard selective approach.
To deal with these problems, Ristani \textit{et al.} proposed the batch-soft triplet loss:
\begin{equation}
\begin{aligned}
L_{3} = \left[ m + \sum_{\mathclap{x_p \in P(a)}} \hspace{1pt} w_p d(x_a,x_p) - \sum_{\mathclap{x_n \in N(a)}} \hspace{1pt} w_n d(x_a,x_n) \right]_+ \\
w_p = \frac{ e^{d(x_a, x_p)} }{ \sum\limits_{\mathclap{\substack{x \in P(a)}}} e^{d(x_a, x)} } \ , \quad w_n = \frac{ e^{-d(x_a, x_n)} }{ \sum\limits_{\mathclap{\substack{x \in N(a)}}} e^{-d(x_a, x)} } \
\label{eq:adaptive_weights}
\end{aligned}
\end{equation}

Observe, that the hyper-parameter $m$, which denotes the margin, exists in all of these triplet loss variations.
Tuning this hyper-parameter manually is not easy, therefore, we next propose an alternative triplet loss that eliminates it.

Our key idea is to replace the hard cutoff max function with an exponential decay $Softplus(\boldsymbol{\cdot}) = \ln(1+exp(\boldsymbol{\cdot}))$ as follows:
\begin{equation}
\begin{aligned}
L_{4} =
Softplus\left(\sum_{{x_p}} \hspace{1pt} w_p d(x_a,x_p)
- \sum_{\mathclap{x_n}} \hspace{1pt} w_n d(x_a,x_n) \right)
\label{eq:final_softtriplet_softmargin}
\end{aligned}
\end{equation}
The soft margin eliminates the margin parameter.

\begin{figure}
\centering
\includegraphics[width=0.4\textwidth]{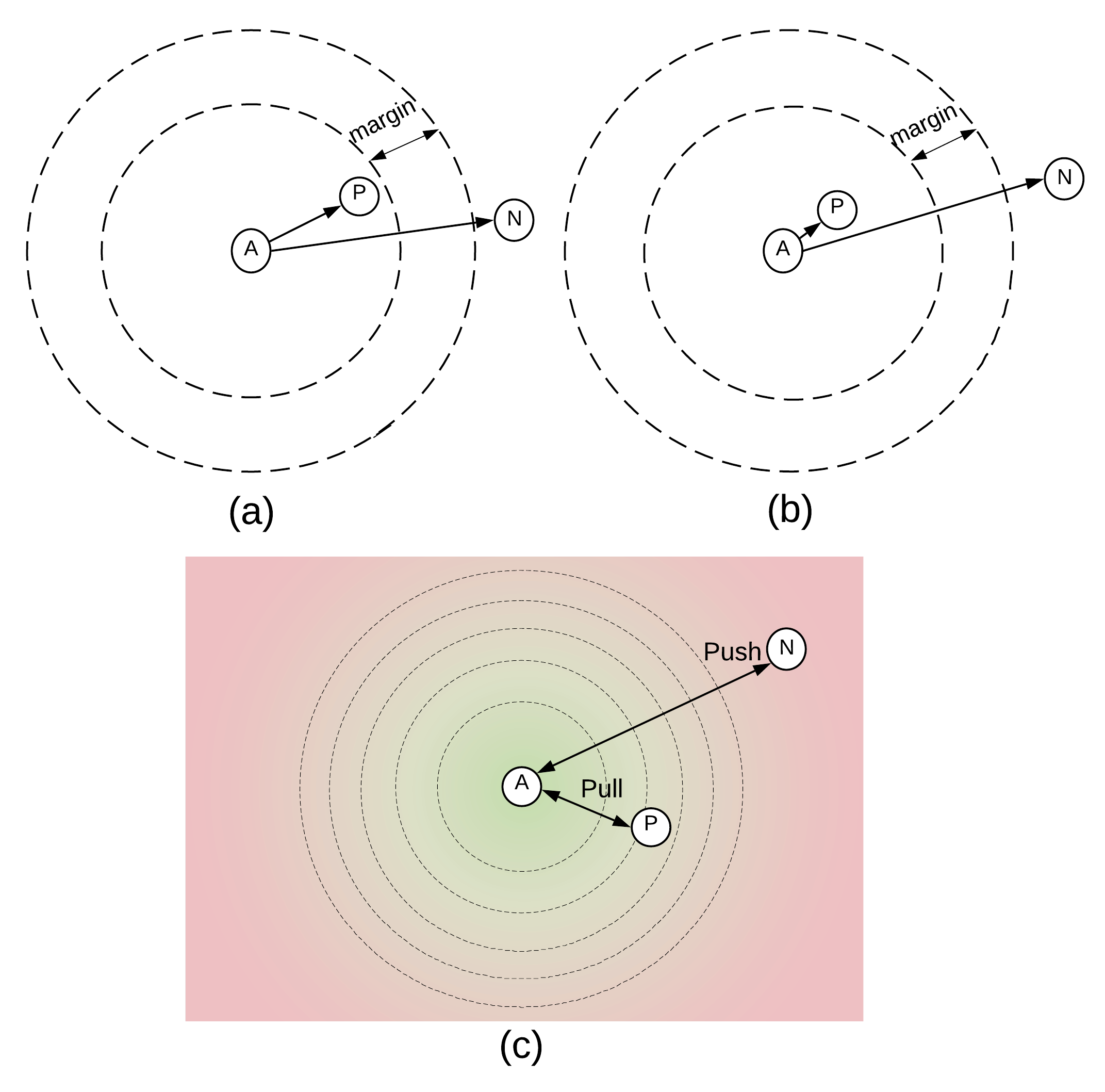}
\caption{Example of hard margin vs soft margin.
The scenario in (b) is more desirable than (a) because the positive sample is closer to the anchor and the negative sample is farther away. This, however, will not be captured by the hard margin triplet loss because both cases correspond to a loss value of zero.
(c) Differently, when using a soft margin the loss will continue to pull the positive sample closer to the anchor while pushing the negative sample away and will encourage going from (a) to (b). }
\label{fig:soft_margin}
\end{figure}

Figure \ref{fig:soft_margin} illustrates one of the benefits of the soft margin over the hard margin.
Using a hard margin value, when the separation between the negative samples and the positive samples becomes larger than the hard margin, the loss is zero and therefore further minimization will not push the positive samples closer or the negative samples farther away from the anchor.
This is illustrated in the examples in (a) and (b) that will both obtain a loss value of zero since both answer the assumption of the hard margin.
Conversely, the soft margin encourages a continuous reduction of the positive distance to the anchor while increasing the negative distance.
This is illustrated in (c), that shows the the computed loss will continue to push the positive sample closer to the anchor while pushing the negative sample away.
\begin{figure}
\centering
\includegraphics[width=0.4\textwidth]{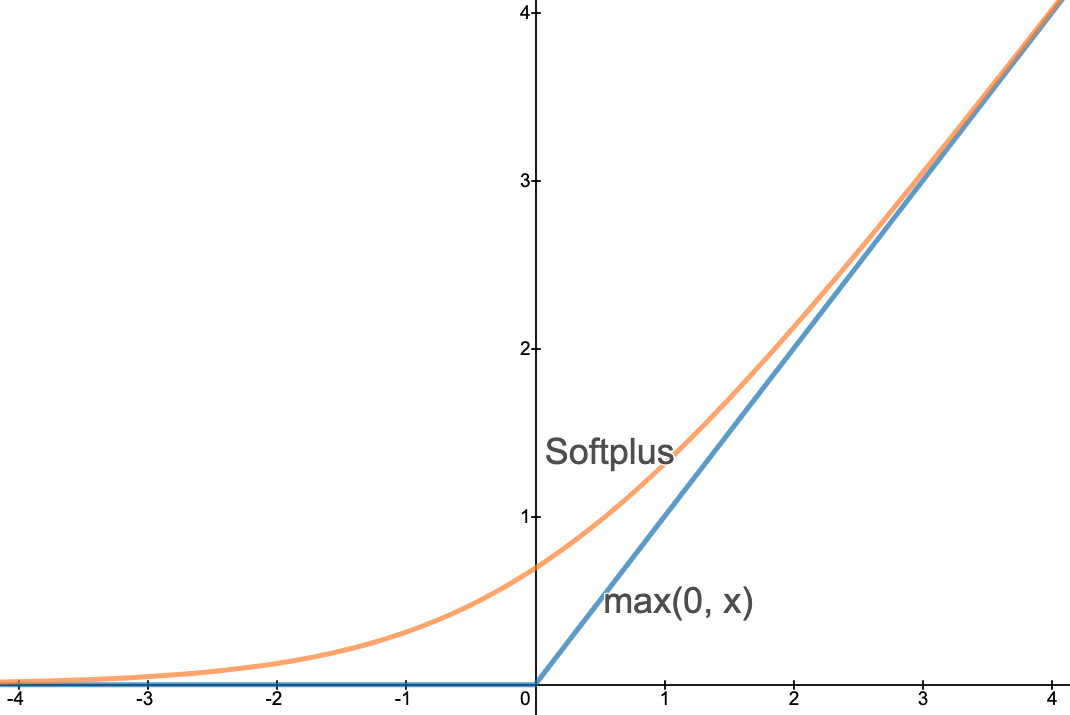}
\caption{The \textit{Softplus} function $(\ln(1+exp(\boldsymbol{\cdot})))$ compared to  $max(0,\boldsymbol{\cdot})$.}
\label{fig:softplus}
\end{figure}

\begin{figure}
\centering
\includegraphics[width=0.42\textwidth]{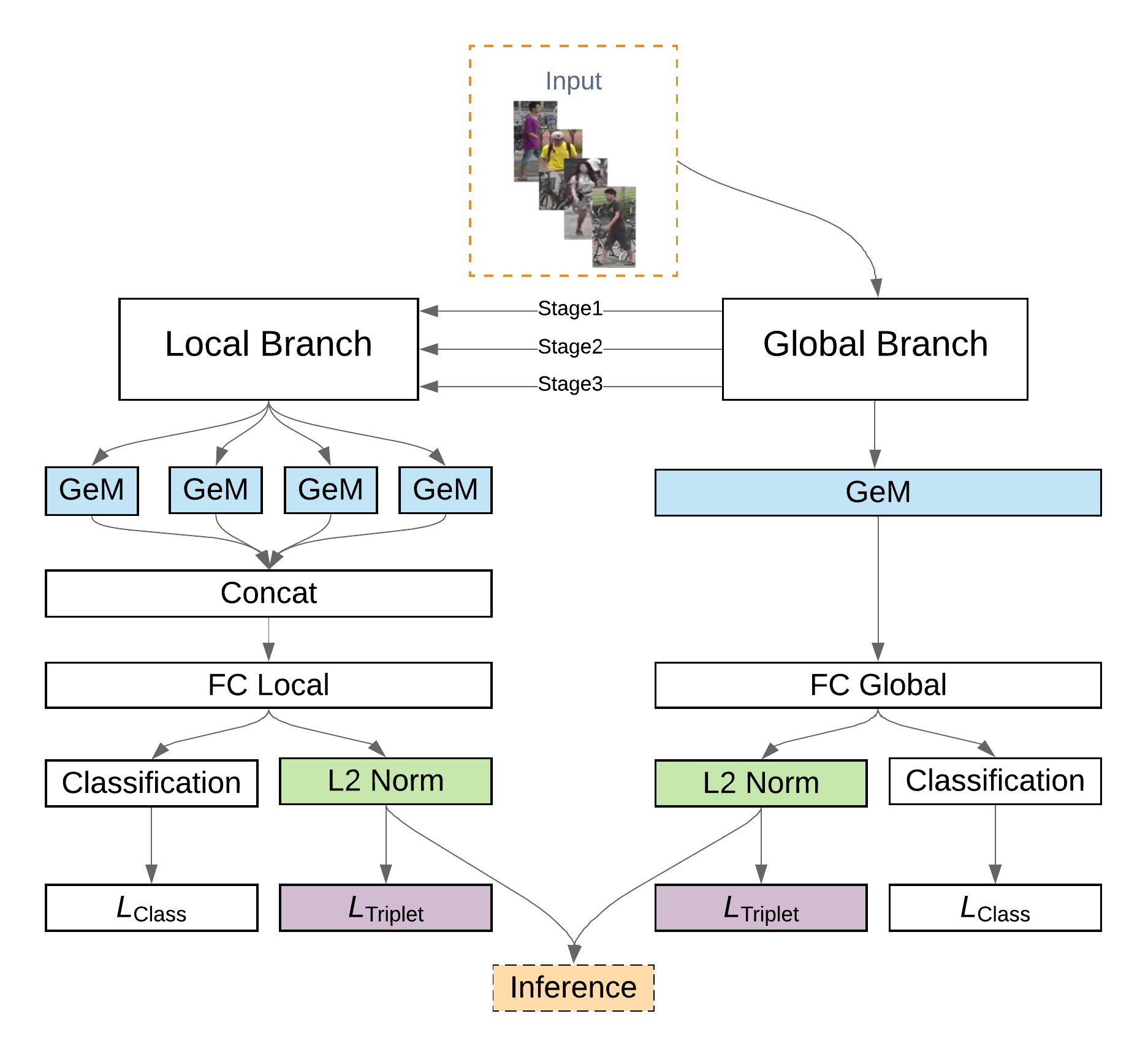}
\caption{Our ReID architecture shows the proposed modifications over the original HA-CNN: $L_2$ normalization during training, GeM instead of average pooling, and soft triplet loss.}
\label{fig:arch_diagram}
\end{figure}

\paragraph{L2 normalization}
The normalization of the feature vectors can be important when using two different loss functions such as cross-entropy and triplet loss which are optimized using different distance measures. \cite{luo2019bag} tackled the normalization problem by adding a batch normalization layer after the feature vectors, right before the fully connected layer. In our empirical studies we found that simply using $L_2$ normalization for each feature vector (global and local) during training achieves an even better performance.
Figure~\ref{fig:arch_diagram} shows the additional $L_2$ normalization used during training and inference.

\paragraph{SWAG \cite{maddox2019simple}}
A common technique to further boost the performance of a model is via ensembles. A common approach is to use an ensemble of models in test time for the final prediction, however, this requires high computing resources.
A more efficient approach is Stochastic weight averaging (SWA) \cite{izmailov2018averaging}, that forms an ensemble during training and outputs a single model for inference. SWA essentially conducts a uniform average over several model weights traversed by SGD during training to achieve a wider region of the loss minima. In order to use SWA a learning rate scheduler is required.

We have made two modifications over HA-CNN ensemble scheme.
First, we follow \cite{maddox2019simple} and use SWA-Gaussian (SWAG). SWAG fits a Gaussian distribution using the SWA solution and diagonal covariance forming an approximate posterior distribution over neural network weights. Next, SWAG performs a Bayesian model averaging based on the Gaussian distribution.
Second, we have found empirically that the original learning rate scheduler of \cite{izmailov2018averaging} can be improved.
We suggest using the cosine annealing learning rate scheduler with $cycles=15$ of 35 epochs and cycle decay factor of 0.7 after each cycle. At the end of each cycle we average the weights of the current model with the previous models taken from the end of each cycle.

\paragraph{Other training techniques}

The random erasing augmentation (REA) \cite{zhong2017random} that randomly erases a rectangle in an image has shown to improve the model generalization ability. We used REA with the following parameters: probability for random erasing an image of $0.5$, area ratio of erasing a region in the range of $0.02<S_e<0.4$, and with aspect ratio in the range of $0.3<r<3.3$.

Warmup \cite{fan2019spherereid} - used to bootstrap the network for better performance. Starting with a smaller learning rate has shown to improve the training process stability, especially when using a randomly initialized model. Using warmup we start the training with a small learning rate and then gradually increase it. We used the following learning rate scheme:
\begin{equation}
lr(t)=\left\{\begin{array}{ll}
{3 \times 10^{-2} \times \frac{t}{10} }  & {\text { if } t \leq 10} \\
{3 \times 10^{-2}} & {\text { if } 10< t \leq 150} \\
{3 \times 10^{-3}}  & {\text { if } 150< t \leq 225} \\
{3 \times 10^{-4}} & {\text { if } 225< t \leq 350}
\end{array}\right.
\label{eq:lr}
\end{equation}

Label smoothing \cite{szegedy2016rethinking} - widely used for classification problems by encouraging the model to be less confident during training and prevent over-fitting. We used label smoothing in a similar way as proposed in \cite{luo2019bag}.

%%%%%%%%%%%%%%%%%%%%%%%%%%%%%%%%%%%%%%%%%%%%%%%%%%%%%%%
\subsection{Architecture modifications}
\label{sec:architecture_modifications}
In addition to the training techniques listed above, we further suggest the following architecture modifications to HA-CNN.

\begin{figure}
\centering
\includegraphics[width=0.48\textwidth]{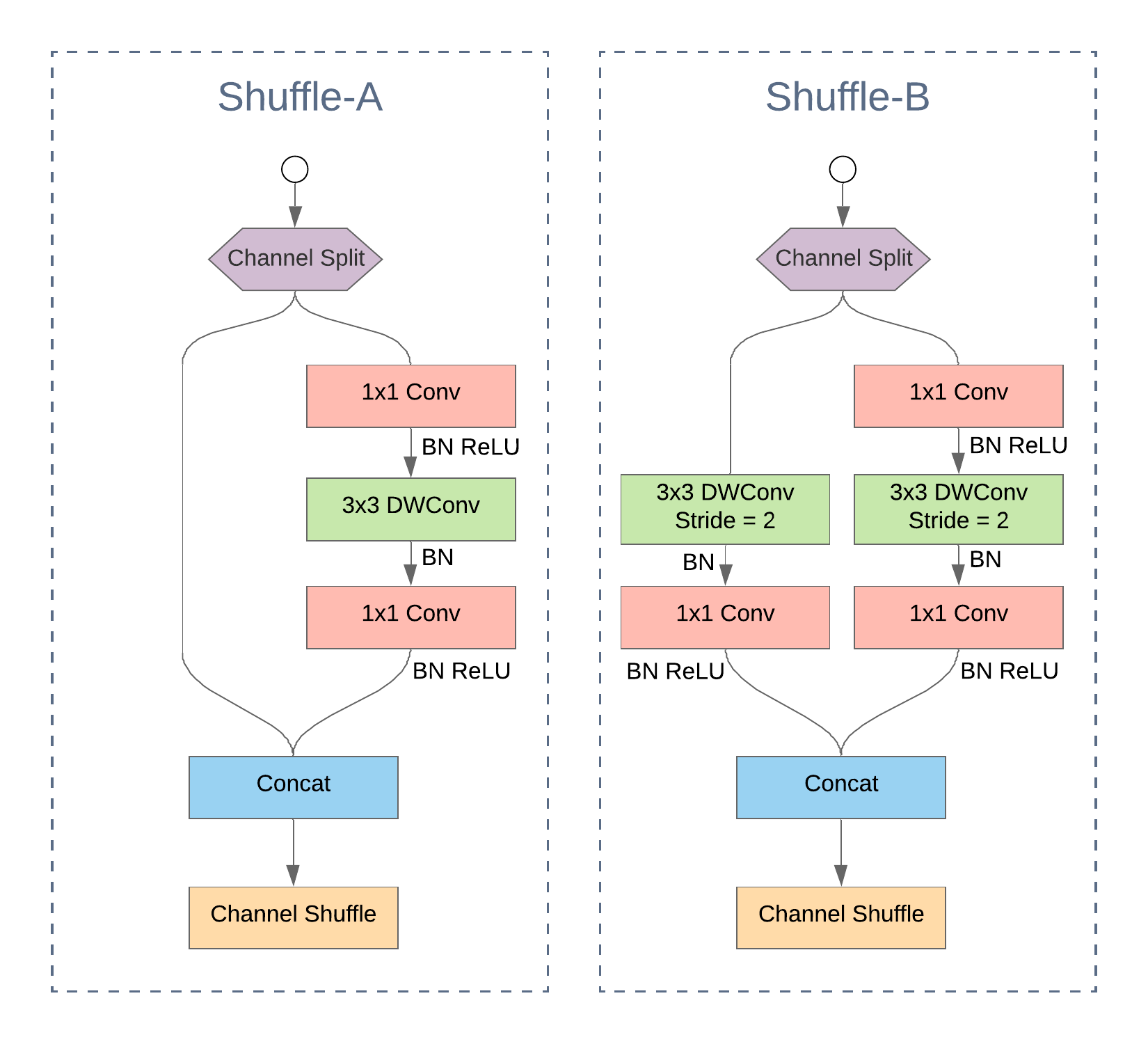}
\caption{The shuffle blocks used in this study to replace the original HA-CNN inception blocks.}
\label{fig:shuffle}
\end{figure}

\paragraph{Shuffle blocks \cite{ma2018shufflenet}}
Our goal was to improve the network accuracy while maintaining a small number of parameters. To do this, we examined replacing the inception blocks with the shuffle blocks presented in Figure~\ref{fig:shuffle}.

Shuffle-A is more efficient than the original inception block since it splits the input features into two equal branches, the first branch remains as is while three convolution operators are applied to the second branch. In addition, one of the convolution operators is depth-wise convolution. The Shuffle-A block can be used in a repeated sequence and still maintain the same number of parameters as the original inception block. Hence, we were able to build a deeper network with a similar number of parameters.
The Shuffle-B block is similar to Shuffle-A but can be used for spatial down-sampling or channel expansion. These characteristics require convolution operators to be applied also to the first branch.
Table \ref{tab:arch} summarizes the repeated sequences of Shuffle blocks used in our proposed architecture.

\begin{table*}
\centering
\begin{tabular}{|l|c|c|c|c|c|c|c|c|}
\hline
\multirow{2}{*}{Local Branch} & \multirow{2}{*}{Global Branch} & \multirow{2}{*}{Layer}                                                     & \multirow{2}{*}{Input}            & \multirow{2}{*}{Stride}                            & \multicolumn{2}{c|}{1$\times$}                                            & \multicolumn{2}{c|}{2$\times$}                                                                 \\
\cline{6-9}
                              &                                &                                                                            &                                   &                                                    & Repeat                                             & Output Ch.           & Repeat                                                      & \multicolumn{1}{l|}{Output Ch.}  \\
\hline
                              & Conv1                          & Conv 3x3                                                                   & 160$\times$64                     & 2                                                  & 1                                                  & 32                   & 1                                                           & \textbf{36 }                     \\
\hline
                              & Stage1                         & \begin{tabular}[c]{@{}c@{}}Shuffle-B\\ Shuffle-A\\ Shuffle-B \end{tabular} & 80$\times$32                      & \begin{tabular}[c]{@{}c@{}}1\\ 1\\ 2 \end{tabular} & \begin{tabular}[c]{@{}c@{}}1\\ 7\\1 \end{tabular}  & \multirow{3}{*}{128} & \begin{tabular}[c]{@{}c@{}}1 \\\textbf{8}\\1 \end{tabular}  & \multirow{3}{*}{\textbf{240}}    \\
\cline{1-6}\cline{8-8}
                              & Soft-Attn1                     & HA-Block                                                                   & 40$\times$16                      & 1                                                  & 1                                                  &                      & 1                                                           &                                  \\
\cline{1-6}\cline{8-8}
Hard-Attn1                   &                                & Shuffle-B                                                                  & 4$\times$(24$\times$28)  & 1                                                  & 1                                                  &                      & 1                                                           &                                  \\
\hline
                              & Stage2                         & \begin{tabular}[c]{@{}c@{}}Shuffle-B\\ Shuffle-A\\ Shuffle-B \end{tabular} & 40$\times$16                      & \begin{tabular}[c]{@{}c@{}}1\\ 1\\ 2 \end{tabular} & \begin{tabular}[c]{@{}c@{}}1\\ 10\\1 \end{tabular} & \multirow{3}{*}{256} & \begin{tabular}[c]{@{}c@{}}1 \\\textbf{11}\\1 \end{tabular} & \multirow{3}{*}{\textbf{320}}    \\
\cline{1-6}\cline{8-8}
                              & Soft-Attn2                     & HA-Block                                                                   & 20$\times$8                       & 1                                                  & 1                                                  &                      & 1                                                           &                                  \\
\cline{1-6}\cline{8-8}
Hard-Attn2                   &                                & Shuffle-B                                                                  & 4$\times$(12$\times$14)  & 1                                                  & 1                                                  &                      & 1                                                           &                                  \\
\hline
                              & Stage3                         & \begin{tabular}[c]{@{}c@{}}Shuffle-B\\ Shuffle-A\\ Shuffle-B \end{tabular} & 20$\times$8                       & \begin{tabular}[c]{@{}c@{}}1\\ 1\\ 2 \end{tabular} & \begin{tabular}[c]{@{}c@{}}1\\ 7\\1 \end{tabular}  & \multirow{5}{*}{384} & \begin{tabular}[c]{@{}c@{}}1 \\\textbf{8}\\1\end{tabular}   & \multirow{5}{*}{\textbf{480}}    \\
\cline{1-6}\cline{8-8}
                              & Soft-Attn3                     & HA-Block                                                                   & 10$\times$4                       & 1                                                  & 1                                                  &                      & 1                                                           &                                  \\
\cline{1-6}\cline{8-8}
Hard-Attn3                   &                                & Shuffle-B                                                                  & 4$\times$(6$\times$7)    & 1                                                  & 1                                                  &                      & 1                                                           &                                  \\
\cline{1-6}\cline{8-8}
                              & Pooling                        & GeM                                                                        & 10$\times$4                       & 1                                                  & 1                                                  &                      & 1                                                           &                                  \\
\cline{1-6}\cline{8-8}
Pooling                       & \multicolumn{1}{l|}{}          & GeM                                                                        & 4$\times$(3$\times$4)    & 1                                                  & 1                                                  &                      & 1                                                           &                                  \\
\hline
                              & FC Global                      & Linear                                                                     & 1$\times$1                        & 1                                                  & 1                                                  & \multirow{2}{*}{512} & 1                                                           & \multirow{2}{*}{\textbf{960}}    \\
\cline{1-6}\cline{8-8}
FC Local                     &                                & Linear                                                                     & 1$\times$1                        & 1                                                  & 1                                                  &                      & 1                                                           &                                  \\
\hline
\hline
FLOPs                         & \multicolumn{4}{c|}{}                                                                                                                                                                                & \multicolumn{2}{c|}{0.72B}                                                & \multicolumn{2}{c|}{\textbf{1.68B }}                                                           \\
\hline
\# of Params.                 & \multicolumn{4}{c|}{}                                                                                                                                                                                & \multicolumn{2}{c|}{2.9M}                                                 & \multicolumn{2}{c|}{\textbf{6.4M }}                                                            \\
\hline

\end{tabular}
\medskip
\caption{Overall architecture of our model, for 2 different levels of complexities. Since our architecture uses a low-resolution input of 160x64, we down-scale the feature maps by applying strided convolution only in the last layer of each stage and not in the beginning. This way the network can leverage a higher spatial resolution in most of the network.}
\label{tab:arch}
\end{table*}

\paragraph{Generalized Mean (GeM) \cite{radenovic2018fine}}
In the original HA-CNN global average pooling was used just before the fully connected layer.
Replacing it with global max pooling gave undecicive results, sometime better and sometimes worse. Therefore, we suggest using the trainable Generalized Mean (GeM) pooling, which generalizes both max and average pooling.
The GeM operator for a single feature map $f_k$ can be written as:
\begin{equation}
\begin{aligned}
GeM(f_k = [x_0,x_1, ..., x_n]) = \left[\frac{1}{n}\sum\limits_{\mathclap{\substack{i = 1 }}}^{n} {x_i}^{p_k} \right]^\frac{1}{p_k}
\label{eq:GeM}
\end{aligned}
\end{equation}
We initialized the parameter $p_k=3$. Figure \ref{fig:arch_diagram} shows where is it used during training and inference.

\paragraph{Deeper and wider}
We further study empirically the impact of using a deeper and wider version of the architecture by modifying the number of shuffle blocks as well as the number of output channels in each stage.
Table~\ref{tab:arch} presents these modifications in bold.

\subsection{Additional tricks we tried}
\label{sec:tried}

For completeness, we list here training options that have been introduced by prior work and our experiments found to deteriorate the results:
\begin{enumerate}
    \item As mentioned before, max and average pooling provide different results so one way to benefit from both pooling methods is by concatenation of their output. Basically we tried to replace the global average pooling used in the original HA-CNN architecture with these two pooling methods and concatenations. It resulted in a similar accuracy with more parameters in the final model.
    \item The batch norm suggested by \cite{luo2019bag} provided inferior results when compared to the simple $L_2$ normalization.
    \item Hard triplet loss instead of the soft version was too sensitive to outliers.
    \item Shuffle blocks without $L_2$ normalization or soft margin in the triplet loss didn't improve the performance.
    \item Training for more epochs didn't improve the performance. The only way it did lead to an improvement was using the Cyclic LR scheme.
    \item Cyclic LR scheme didn't improve the results when used from scratch from the beginning of the training. It only worked when used in additional training epochs after the the model converged.
\end{enumerate}

\section{Experimental results}
\label{sec:experiments}
In the following we evaluate our models on Market1501 and DukeMTMC ReID datasets based on rank-1 accuracy and mAP. Next, the performance boost by each methods presented in section \ref{sec:tricks} is evaluated.

\paragraph{Implementation details}
All person images are resized to $160 \times 64$. We used SGD for optimization with a linear warm-up as in Equation~\eqref{eq:lr} for a total of 350 epochs. When using SWAG we train for 15 cycles of 35 epochs which sums up to 525 additional epochs. We randomly sample 8 identities and 4 images per person in each training batch.

\renewcommand{\multirowsetup}{\centering}
\begin{table}[tb]\footnotesize
  \begin{center}
  \begin{tabular}{ ccc|cc|cc}
\hline
    		&		& \multicolumn{2}{c|}{Market1501} & \multicolumn{2}{c}{DukeMTMC}	 \\
  Type   & Method & r = 1 	& mAP	&r = 1 	& mAP 	 \\
 	\hline
% 	\hline
%     \multirow{3}{1cm}{Pose-guided}& GLAD\cite{wei2017glad}       &89.9	&73.9	&-	&-		\\
%                                 & PIE \cite{zheng2019pose}      &87.7	&69.0	&79.8	&62.0		\\
%                                 & PSE \cite{saquib2018pose}     &78.7	&56.0	&-	&-		\\
    \hline
    \multirow{2}{1cm}{Mask-guided}& SPReID \cite{kalayeh2018human}   & 92.5 & 81.3	& 84.4	&71.0		\\
                                & MaskReID \cite{qi2018maskreid}     &90.0	&75.3	&78.8	&61.9		\\
 \hline
    \multirow{9}{1cm}{Stripe-based}& AlignedReID \cite{zhang2017alignedreid}&90.6 &77.7 &81.2	&67.4 		\\
                                & SCPNet \cite{fan2018scpnet}       & 91.2	&75.2	&80.3	&62.6		\\
                                & LocalCNN \cite{yang2018local}    & 91.5	& 77.7	&82.2	&66.0		\\
                                & Pyramid\cite{zheng2018pyramid}     & 92.8 &82.1	&-	&-		\\
                                & PCB \cite{sun2018pcb}          & 93.8	&81.6	&83.3	&69.2		\\
                                & BFE\cite{dai2018batch}            & 94.5 &85.0	&88.7	&75.8		\\
                                & MGN \cite{wang2018learning}  & 95.7	& 86.9	&88.7	&78.4		\\
                                & Pyramid\cite{zheng2018pyramid}     & 95.7 &88.2	&89.0	&79.0		\\
                                & LocalCNN (MG) \cite{yang2018local}  & 95.9	& 87.4	&-	&-		\\
\hline

    	\multirow{2}{1cm}{Dense-semantics}& \multirow{2}{1.2cm}{DSA \cite{zhang2019densely}}  & \multirow{2}{0.5cm}{95.7}	& \multirow{2}{0.5cm}{87.6}	&\multirow{2}{0.5cm}{86.2}	&\multirow{2}{0.5cm}{74.3}		\\ & & & & & \\
\hline
    \multirow{3}{1cm}{GAN-based}& Camstyle \cite{zhong2018camstyle} &88.1 	&68.7	&75.3	&53.5		\\
                                & PN-GAN \cite{qian2018pose}        &89.4 	&72.6	&73.6	&53.2		\\
                                & DG-Net \cite{zheng2019joint}        &94.8 	&86.0	&86.6 	&74.8		\\
    \hline
    \multirow{6}{1cm}{Global feature}& IDE \cite{zheng2018discriminatively}  & 79.5	& 59.9	& -	&-		\\
                                & SVDNet \cite{sun2017svdnet}   & 82.3	& 62.1	& 76.7	&56.8		\\
                                & TriNet\cite{hermans2017defense}  & 84.9	& 69.1	& -	& -		\\
                                & AWTL\cite{ristani2018features}  & 89.5	& 75.7	& 79.8	& 63.4		\\
                                & OS-Net \cite{zhou2019omni}    &94.8	&84.9	&88.6	&73.5		\\
                                & BagOfTricks \cite{luo2019bag}    &94.5	&85.9	&86.4	&76.4		\\
   \hline
    \multirow{1}{1cm}{NAS}& Auto-ReID \cite{quan2019auto}  & 94.5	& 85.1	& 88.5	& 75.1		\\
   \hline
    \multirow{6}{1cm}{Attention-based}& HA-CNN \cite{li2018harmonious}   & 91.2	& 75.7	&80.5	&63.8		\\
                                & DuATM \cite{si2018dual}          & 91.4	& 76.6	&81.2	&62.3		\\
                                & Mancs \cite{wang2018mancs} &93.1   &82.3	&84.9	&71.8		\\
                                & ABD \cite{chen2019abd} &95.6   &88.3	& 89.0	&78.6		\\
                                & RGA-SC \cite{zhang2019relation} &95.8   &88.1	& 86.1	&74.9		\\
                                 & \textbf{Ours (2.9M)}   & 95.8	& 88.7	&88.8	&78.9		\\
                                 & \textbf{Ours (6.4M)}   & \textbf{96.2}	& \textbf{89.7}	&\textbf{89.8}	&\textbf{80.3}		\\
    \hline

  \end{tabular}
  \end{center}
  \caption{\label{tab:sota}Comparison of state-or-the-arts methods.}
\end{table}

\begin{table*}
	\centering
	\setlength{\tabcolsep}{3pt}
	\renewcommand{\arraystretch}{0.95}
	\begin{tabular}{r|c|ccccccccc|c}
    	 & HA-CNN \cite{li2018harmonious} & a & b  & c & d & e & f & g & h & i & Compact-ReID \\
    	\hline
        BagOfTricks \cite{luo2019bag}& & \checkmark & \checkmark & \checkmark & \checkmark & \checkmark &\checkmark &\checkmark & \checkmark & \checkmark & \checkmark\\
        \hline
        Soft triplet &  & &  & \checkmark & \checkmark & \checkmark &\checkmark &\checkmark & \checkmark & \checkmark & \checkmark\\
        L2 normalization  &  &  & \checkmark &  & \checkmark & \checkmark &\checkmark &\checkmark & \checkmark & \checkmark & \checkmark\\
        Shuffle blocks     &  & & \checkmark & \checkmark &  & \checkmark &\checkmark &\checkmark & \checkmark & \checkmark & \checkmark\\
        Soft margin  &  & & & \checkmark & \checkmark &  &\checkmark &\checkmark & \checkmark & \checkmark & \checkmark\\
        GeM                & & & \checkmark & \checkmark & \checkmark & \checkmark &  &\checkmark & \checkmark & \checkmark & \checkmark\\
        Deeper  \& wider  &  & & \checkmark & \checkmark & \checkmark & \checkmark &\checkmark &  &  & \checkmark & \checkmark\\
        SWAG              & & & & & & & & &\checkmark &  &\checkmark \\
        \hline
        \multirow{ 2}{*}{Market1501} Rank1
                          & 91.2 & 93.2 & 94.1 & 94.5 & 94.9 & 95.1 & 95.3 & 95.4 & 95.8  & 95.7 & 96.2 \\
        mAP               & 75.7 & 82.0 & 85.3 & 85.7 & 86.6 & 87.1 & 87.9 & 87.3 & 88.7  & 88.1 & 89.7\\

	\end{tabular}
    \medskip
    \caption{Ablation study on Market1501. The first column indicates the different training techniques and architecture modifications we tried including some of the tricks mentioned in BagOfTricks \protect\cite{luo2019bag}: warmup, random erase, label smoothing, and no bias in the classification layers. The baseline we started with, i.e. the original HA-CNN implementation, is presented in the second column for comparison. The last column shows the results of our proposed Compact-ReID network including all of the training techniques and architecture modifications proposed in this study. Columns a-i demonstrates the impact of each modification by turning it off.}
    \label{tab:small_ablation}
\end{table*}

\subsection{Comparison to state of the arts}
We compare our models performance to several state of the art methods (Table \ref{tab:sota}).
Our best model achieves state of the art results in terms of rank-1 accuracy and mAP on Market1501 (96.2, 89.7) and DukeMTMC (89.8, 80.3) with only 6.4M parameters.
To our best knowledge, our model achieves the best performance on these public datasets. It should be noted that the smaller version of our model (2.9M parameters) also achieves state of the art results on both datasets.

In terms of FLOPS our final network has 1.7B FLOPS while the ResNet 50 used in Luo \textit{et al.} \cite{luo2019bag} implementation has 4.1B FLOPS. We did not apply re-ranking for clear comparison and since it is currently not relevant for real world practice.

\subsection{Ablation study}
To evaluate the different training techniques explored in this study we set several experiments in an ablation study. Table \ref{tab:small_ablation} shows the different modifications starting from the original HA-CNN architecture. The first row indicates using some of the tricks from \cite{luo2019bag} that showed an improvement when tested on Market1501 using the HA-CNN architecture. These include warm-up, random erasing, and no-bias in the fully connected layers. These tricks alone (experiment a) provided an improvement of 2\% in rank-1 accuracy and 6.3\% in mean average precision compared to the original HA-CNN paper result (i.e. our baseline).

Next, to test the influence of some of our modifications we report the performance after disabling them. The most significant decrease in results compared to column i was caused by disabling the weighted triplet loss and soft margin (using the original triplet loss as in equation \eqref{eq:triplet_general} instead) with a drop of 1.6\% in rank-1 accuracy and 2.8\% in mAP (column b). Cancelling the $L_2$ normalization caused a decrease of 1.2\% in rank-1 accuracy and 2.4\% in mAP (column c). Reduction of other modifications such as shuffle blocks, soft margin, GeM, and deeper and wider network caused a decrease in the performance as well indicating the benefit of using it.

Finally, we used the SWAG in two experiments: experiment h and the final Compact-ReID. Continuing the training with SWAG provided an improvement in both rank-1 and mAP in both experiments. The SWAG is used in this study as a post process for models that already achieve high accuracy to show its contribution on top of that.

\begin{table}\small
  \begin{center}
  \begin{tabular}{c|c|cc}
    	\multirow{2}{*}{Setup}	& \multirow{2}{*}{Update}	& \multicolumn{2}{c}{Market1501} 	 \\
   &  	& r = 1 	& mAP 	    \\
 	\hline
	\hline
\multirow{3}{*}{1} & - &93.8	&83.6		\\
& +LR Scheme  & 94.3 & 84.8       \\
& +SWAG &\textbf{94.5}	&\textbf{85.3}		\\
 	\hline
\multirow{3}{*}{2} & - &95.4	&87.3		\\
 & +LR Scheme &\textbf{95.8}& 88.2       \\
 & +SWAG &\textbf{95.8}	&\textbf{88.7}		\\
 	\hline
\multirow{3}{*}{3} & - &95.7	&88.1		\\
 & +LR Scheme  & 95.7 & 88.9       \\
 & +SWAG &\textbf{96.2}	&\textbf{89.7}		\\
\hline

  \end{tabular}
  \end{center}
    \caption{Performance evaluation on Market1501 for SWAG with cosine annealing with decay factor learning scheme.}
    \label{tab:SWAG}
\end{table}

\subsection{Exploring SWAG}
\label{sec:SWAG}
Our empirical experiments showed that the SWAG method consistently improved our model performance. However, it requires additional training time and uses a custom made cosine annealing learning scheme with a decay factor. Therefore, we wanted to further explore the SWAG contribution by analyzing some of our experimental results. Table \ref{tab:SWAG} shows the results when testing the learning rate scheme with and without SWAG for three different setups. In the first setup we used our proposed architecture minus three main modifications: GeM, Shuffle blocks, and deeper and wider. The second and third setups are experiments g and i in Table \ref{tab:small_ablation} respectively. Evidently, adding the LR scheme provided a nice improvement, and adding the SWAG performed even better. The most significant improvements were in terms of mAP.

\begin{figure}
\label{fig:swag_market}
\centering
\begin{minipage}[b]{0.4\textwidth}
\includegraphics[width=1\textwidth]{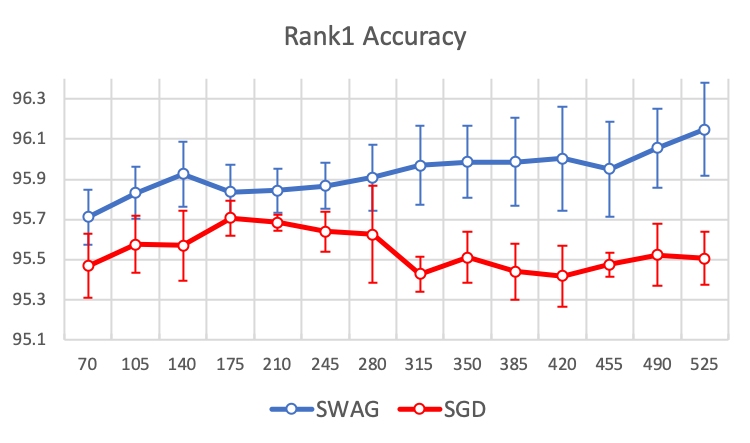}
\end{minipage}

\centering
\begin{minipage}[b]{0.4\textwidth}
\includegraphics[width=1\textwidth]{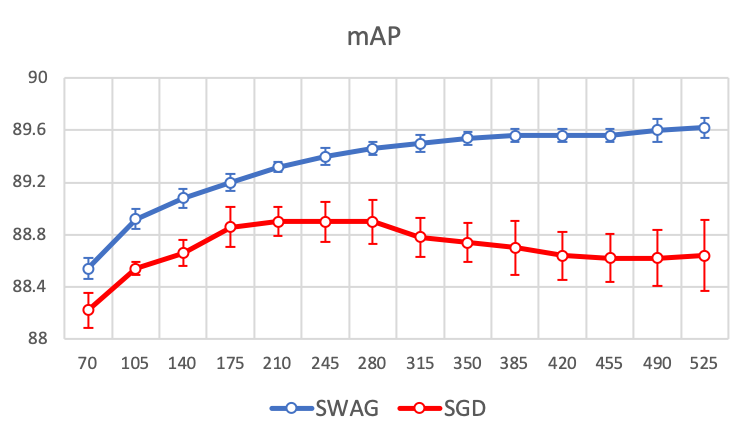}
\end{minipage}
\caption{Performance evaluation of SWAG compared to SGD using the cosine annealing learning scheme on Market1501 dataset showing the average of 5 runs. Top: rank-1 accuracy vs. epoch. Bottom: mAP vs. epoch.}\label{fig:SWAG}
\end{figure}

Figure~\ref{fig:SWAG} presents the average over five experiments comparing SWAG and standard SGD in terms of Rank1 accuracy and mAP on Market1501 dataset. Using SWAG the accuracy trend seems more consistent compared to standard SGD. In addition, it is significantly better in terms of mAP.

\section{Application to multi object tracking}
\label{sec:tracking}
Although the public datasets used in this study for person ReID are valuable for comparison between different architectures and models, we wanted to evaluate the model's applicability by using it to improve multi target multi camera tracking. Testing the model in a real world setting such as tracking is much more challenging. A wrong ReID assignment can affect the assignment of other persons since we only compare each query image to tracks that are not active (not present in the room at the time of the query). In addition, for each query we need to decide if we open a new track or assign it to an existing track (ReID), meaning that in some cases the gallery does not include images of the person found in the query.

We used the LAB sequence which is a part of the Task-Decomposition database \cite{hu2014dynamic} of multi-view sequences for people tracking. The LAB sequence is about 12.5 minutes long\footnote{Information in the database website mentions 3.5 minutes but the downloaded videos are actually 12.5 minutes long.}, the tracking domain is about 5*6 meters in dimension, and the images were captured at 15 Hz with a resolution of 640*480 pixels, where four cameras are installed at the corners of the room. Through the sequence, people enter, walk around, sit down and exit the room randomly, causing frequent occlusions. The maximum number of people in the scene at the same time is 7.
We first used an internal software for global people tracking which uses the calibration provided for each camera and report the results we got with and without using the model for ReID in terms of MOTA and IDF1. We used ReID each time a person enters the room by comparing it to several images per person that is currently not tracked inside the room.

Table~\ref{tab:tracking_benchmark_indoor} shows the results obtained using different models including: the original HA-CNN and our proposed model. Our model performed better than the original HA-CNN in terms of IDF1 using Market1501 or DukeMTMC for training. Due to the original resolution of the videos, the size of the bounding box of each query and gallery image can get very small in size. Our model showed robustness to the low-res images since it was trained on small sized input.

\begin{table}
\centering

\begin{tabular}{c|c|c|c}
Model       & Trained Dataset & MOTA & IDF1  \\
\hline\hline
Compact-ReID & DukeMTMC        &   96.1   &   \textbf{89.1}    \\
Compact-ReID & Market1501      &   96.1   &   79.6   \\
HA-CNN      & DukeMTMC        &   96.1   &   78.9   \\
HA-CNN      & Market1501      &   96.1   &   65.7   \\
No ReID     &    -            &   96.1   &   57.1
\end{tabular}
\medskip
\caption{Multi camera multi target tracking results on LAB dataset using our proposed Compact-ReID model compared to the original HA-CNN.}
\label{tab:tracking_benchmark_indoor}
\end{table}

\section{Conclusions}
This paper explores several training techniques and architecture modifications focusing on a small-sized randomly initialized attention network for person ReID. Each training technique is tested as well as some of the tricks presented in other prior works. Using the proposed training scheme and network modifications we were able to outperform SotA works achieving 96.2\% rank1 accuracy and 89.7\% mAP on Market1501 and 89.8\% rank1 accuracy and 80.3\% mAP on DukeMTMC with only 6.4M parameters. In addition, we show that even for a smaller version (2.9M parameters) we achieve state of the art results. Finally, we show the applicability of our proposed model by utilizing it to improve existing methods for multi object tracking on a public dataset.
Future work entails more experiments using other deep ReID networks as our baseline, as well as tackling the cross-domain challenges in person ReID.

\begin{acks}
We would like to thank Sagi Rorlich and Genadiy Vasserman for their help in some of the experiments.
\end{acks}

  \bibliographystyle{ACM-Reference-Format}
  \balance
  \bibliography{egbib}

\end{document}